\documentclass[english]{elsarticle}
\usepackage[T1]{fontenc}
\usepackage[latin9]{inputenc}
\usepackage{amsmath}
\usepackage{amssymb}
\usepackage{graphicx}

\makeatletter

\providecommand{\tabularnewline}{\\}

\usepackage{multicol}

\makeatother

\usepackage{babel}
\begin{document}

\title{A dense subgraph based algorithm for compact salient image region
detection}

\author{\begin{multicols}{2}  Souradeep Chakraborty*\\Department of Electronics and Electrical Communication Engineering,\\ Indian Institute of Technology,\\ Kharagpur 721302, India\\ Email : sourachakra@gmail.com\\
Pabitra Mitra\\Department of Computer Science and Engineering,\\ Indian Institute of Technology,\\ Kharagpur 721302, India\\ Email : pabitra@gmail.com \end{multicols}}
\begin{abstract}
We present an algorithm for graph based saliency computation that
utilizes the underlying dense subgraphs in finding visually salient
regions in an image. To compute the salient regions, the model first
obtains a saliency map using random walks on a Markov chain. Next,
$k$-dense subgraphs are detected to further enhance the salient regions
in the image. Dense subgraphs convey more information about local
graph structure than simple centrality measures. To generate the Markov
chain, intensity and color features of an image in addition to region
compactness is used. For evaluating the proposed model, we do extensive
experiments on benchmark image data sets. The proposed method performs
comparable to well-known algorithms in salient region detection.\end{abstract}
\begin{keyword}
Visual saliency, Markov chain, Equilibrium distribution, Random walk,
$k$-dense subgraph, Compactness
\end{keyword}
\maketitle

\section{Introduction}

The saliency value of a pixel in an image is an indicator of its distinctiveness
from its neighbors and thus its ability to attract attention. Visual
attention has been successfully applied to many computer vision applications,
e.g. adaptive image compression based on region-of-interest {[}35{]},
object recognition {[}36,37,38{]}, and scene classification {[}39{]}.
Nevertheless, salient region and object detection still remains a
challenging task.

The goal of this work is to extract the salient regions in an image
by combining superpixel segmentation and a graph theoretic saliency
computation. Dense subgraph structures are exploited to obtain an
enhanced saliency map. First, we segment the image into regions or
superpixels using the SLIC (Simple Linear Iterative Clustering) superpixel
segmentation method {[}19{]}. Then we obtain a saliency map using
the graph based Markov chain random walk model proposed earlier {[}1{]},
considering intensity, color and compactness as features. Using this
saliency map, we create another sparser graph, on which $k$-dense
subgraph is computed. We get a refined saliency map by this technique.
The use of dense subgraph on a graph constructed from segmented image
regions helps to filter out the densely salient image regions from
saliency maps.

A number of graph based saliency algorithms are known in literature
that improvise on the dissimilarity measure to model the inter-nodal
transition probability {[}1{]}, provide better random walk procedures
like random walk with restart {[}18{]}, or use combinations of different
functions of transition probabilities e.g., site rate entropy function
{[}14{]} to build their saliency model. However, most graph based
methods produce a blurred saliency map. It would be useful to postprocess
the map to filter out the most salient portions. Our model also being
a graph based method, uses dense subgraph computation to filter out
salient regions after random walk is employed. The suppression of
non-salient regions combined with salient region shape retention,
yields saliency maps more closely resembling ground truth data as
compared to the existing methods used for comparison. This has been
achieved by considering a more informative local graph structure,
namely, dense subgraphs, than simple centrality measures in obtaining
the map.

The remainder of the paper is organized as follows. Section 2 surveys
some previously proposed saliency detection algorithms. Section 3
describes the proposed saliency detection procedures. Section 4 is
devoted to experimental results and evaluation. We demonstrate that
the proposed method achieves superior performance when compared to
well-known models, on standard image data sets and also preserves
the overall shapes and details of salient regions quite reliably.
Finally in section 5, we conclude this paper with future research
issues.

\section{Related Work}

Saliency computation is rooted in psychological theories about human
attention, such as the feature integration theory (FIT) {[}44{]}.
The theory states that, several features are processed in parallel
in different areas of the human brain, and the feature locations are
collected in one \textquotedblleft{}master map of locations\textquotedblright{}.
From this map, \textquotedblleft{}attention\textquotedblright{} selects
the current region of interest. This map is similar to what is nowadays
called \textquotedblleft{}saliency map\textquotedblright{}, and there
is strong evidence that such a map exists in the brain. Inspired by
the biologically plausible architecture proposed by Koch and Ullman
{[}45{]}, mainly designed to simulate eye movements, Itti et al. {[}3{]}
introduced a conceptually computational model for visual attention
detection. It was based on multiple biological feature maps generated
by mimicking human visual cortex neurons. It is related to the FIT
theory {[}44{]} which outlines the human visual search strategies.
The recent survey by Borji and Itti {[}47{]} lists a series of visual
attention models {[}47{]}, which demonstrate that eye-movements are
guided by both bottom-up (stimulus-driven) and top-down (task-driven)
factors.

Region based saliency models have been proposed in a number of works.
The early work of Itti \textit{et al}. {[}3{]} was extended by Walther
and Koch {[}15{]} who proposed a way to extract proto-objects. Proto-objects
are defined as spatial extension of the peaks of this saliency map.
This approach calculates the most salient points according to the
spatial-based model, henceforth the saliency is spread to the regions
around them. The work in {[}31{]} addresses the problem of detecting
irregularities in visual data, e.g., detecting suspicious behaviors
in video sequences, or identifying salient patterns in images. The
problem is posed as an inference process in a probabilistic graphical
model. The framework in {[}1{]}, is a computer vision implementation
of the object based attention model of {[}6{]}. In this paper, a grouping
is done by conducting a segmentation method, which acts as the operation
unit for saliency computation. In {[}7{]}, the problem of feature
map generation for region-based attention is discussed, but a complete
saliency model has not been proposed. The method in {[}8{]} is based
on preattentive segmentation, dividing the image into segments, which
serve as candidates for attention, and a stochastic model is used
to estimate saliency. In {[}9{]}, visual saliency is estimated based
on the principle of global contrast, where the region is employed
in computation and is used primarily for the sake of speed up. In
the work {[}23{]}, two characteristics: rareness and compactness have
been utilized. In this approach, rare and unique parts of an image
are identified, followed by aggregating the surrounding regions of
the spots to find the salient regions thus imparting compactness to
objects. In the method followed in {[}22{]}, saliency is detected
by over-segmenting an image and analyzing the color compactness in
the image. Li \textit{et al.}, in their work {[}33{]}, offer two contributions.
First, they compose an eye movement dataset using annotated images
from the PASCAL dataset {[}54{]}. Second, they propose a model that
decouples the salient object detection problem into two processes:
1) a segment generation process, followed by 2) a saliency scoring
mechanism using fixation prediction. A novel propagation mechanism,
dependent on Cellular Automata, is presented in {[}34{]} which exploits
the intrinsic relevance of similar regions through interactions with
neighbors. Here, multiple saliency maps are integrated in a Bayesian
framework.

Several graph based saliency models have been suggested so far. It
is shown in {[}10{]} that gaze shift can be considered as a random
walk over a saliency field. In {[}11{]}, random walks on graphs enable
the identification of salient regions by determining the frequency
of visits to each node at equilibrium. Harel \textit{et al}. {[}1{]}
proposed an improved dissimilarity measure to model the transition
probability between two nodes. These kinds of methods consider information
to be the driving force behind attentive sampling and use feature
rarity to measure visual saliency. In this article, we base our model
on such a graph based model and utilize the embedded dense subgraphs
to better extract the most salient regions from an image. The work
in {[}12{]} provides a better scheme to define the transition probabilities
among the graph nodes and thus constructs a practical framework for
saliency computation. Wang \textit{et al}. {[}14{]} generated several
feature maps by filtering an input image with sparse coding basis
functions. Then they computed the overall saliency map by multiplying
saliency maps obtained using two methods: one is the random walk method
and the other based on the entropy rate of the Markov chain. Gopalakrishnan
\textit{et al.} {[}12{]}, {[}13{]} formulated the salient region detection
as random walks on a fully connected graph and a $k$-regular graph
to consider both global and local image properties in saliency detection.
They select the most important node and background nodes and used
them to extract a salient object. Jiang \textit{et al.} {[}30{]} consider
the absorption time of the absorbing nodes in a Markov chain (constructed
on a region similarity graph) and separate the salient objects from
the background by a global similarity measure. Yang \textit{et al.}
{[}29{]} ranks the similarity of the image regions with foreground
cues or background cues via graph-based manifold ranking, and detects
background region and foreground salient objects. In a more recent
work by {[}32{]}, a novel bottom-up saliency detection approach has
been proposed that takes advantage of both region-based features and
image details. The image boundary selection is optimized by the proposed
erroneous boundary removal and regularized random walks ranking is
implemented to formulate pixel-wised saliency maps from the superpixel-based
background and foreground saliency estimations. 

Many other saliency systems have also been presented in previous years.
There are approaches that are based on the spectral analysis of images
{[}42, 46{]}, models that base on information theory {[}40, 41{]},
Bayesian theory {[}50, 51{]}. Other algorithms use machine learning
techniques to learn a combination of features {[}48,49{]} or employ
deep learning techniques {[}43{]} to detect salient objects.

\section{Proposed Saliency Model}

\begin{figure}
\hspace{1.5cm}\includegraphics[width=9.9cm]{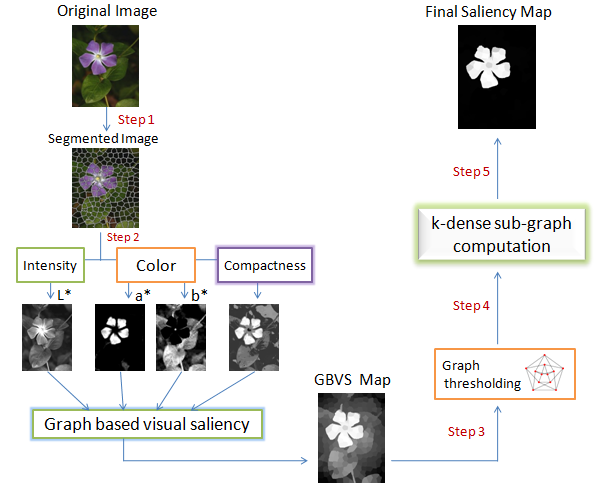}

\caption{\label{fig:1}Flowchart for the proposed saliency computation model.}
\end{figure}

Our method aims to enhance graph based saliency computation techniques
by considering higher level graph structures as compared to those
utilized in Markov chain based measures. Note that it can be used
in conjunction with any graph based saliency computation algorithm.
We use superpixels, that are obtained by pre-segmentation while constructing
the graphs. The goal here is to extract salient regions rather than
pixels. In this section, we describe the proposed model of region-based
visual saliency. We follow a multi-step approach to saliency detection.
The block diagram is illustrated in Figure \ref{fig:1} and the steps
are mentioned below:

\textit{Step 1}: SLIC superpixel segmentation method {[}19{]} is applied
on the original image to generate image regions or superpixels.

\textit{Step 2}: A saliency map is obtained by implementing graph
based saliency model {[}1{]} on the region based graph taking three
feature channels $L^{*}$, $a^{*}$ and $b^{*}$ (considered from
the CIEL{*}a{*}b{*} color space) and the compactness factor for saliency
computation.

\textit{Step 3}: The graph corresponding to the saliency map obtained
in \textit{step 2} is edge thresholded to form a sparser graph.

\textit{Step 4}: Dense subgraph computation is performed on the sparse
graph constructed in \textit{step 3}, which results in detection of
highly salient regions.

\textit{Step 5}: Final saliency map is obtained after saliency assignment
based on \textit{step 4} followed by map normalization.

It might be observed from Figure \ref{fig:1}, that we extract the
feature information and then apply the graph based saliency model
to get an intermediate saliency map, which is further refined by dense
subgraph computation to obtain the final saliency map. The method
constructs the connectivity graph based on image segments or superpixels,
unlike Harel \textit{et al.} {[}1{]} which computes the graph based
on rectangular regions. Throughout the paper, CIEL{*}a{*}b{*} color
space has been used, as Euclidean distances in this color space are
perceptually uniform and it has been experimentally found out in {[}17{]}
to give better results as compared to HSV, RGB and YCbCr spaces. We
describe in subsequent sections the individual steps in details.

\subsection{Superpixel Segmentation and Feature Extraction}

\textit{Superpixel Segmentation}: The image is segmented using SLIC
superpixel segmentation {[}19{]}. Firstly, the RGB color image is
converted to the CIE L{*}a{*}b{*}, a perceptual uniform color space,
which is designed to approximate human vision. The next step consists
of creating superpixels using SLIC algorithm which divides the image
into smaller regions. A value of 250 pixels per superpixel is used
in our experiment. For higher values of pixels per superpixel, the
computation time increases and for lower values of it, region boundaries
are not preserved well. 

\medskip{}

\textit{Feature Extraction}: Four feature channels in three different
spatial scales ($1$, $\frac{1}{2}$ and $\frac{1}{4}$) of the image
are extracted. As the L{*} channel (a measure of lightness) relates
to the intensity of an image, it is considered a feature channel.
Similarly, as a{*} and b{*} components of the CIEL{*}a{*}b{*} color
space correspond to the opponent colors, they are taken as feature
channels representing the color aspect of the image. The fourth feature
channel represents the compactness aspect of regions in the image.
Normalized maps of the 12 ($=4\times3$) feature channels are used
in this experiment. All maps in this paper are normalized as per Equation
\ref{eq:1}. Note here that, we do not use the multi-angle gabor filter
based orientation maps, unlike {[}1{]}. We rather incorporate compactness
as a feature, as most salient objects tend to have compact image regions
as well as well defined object boundaries and the compactness measure
ensures that the background regions with relatively less compact regions,
receive lesser mean region saliency values in further computations.
The red-green and the yellow-blue opponent colors feature (used in
{[}1{]}) are represented by the a{*} and the b{*} channels respectively. 

\begin{equation}
NormMap(M_{i})=\frac{M_{i}-M_{min}}{M_{max}-M_{min}},\label{eq:1}
\end{equation}

\noindent where $M_{i}$ is the feature map value at pixel $i$. $NormMap(M_{i})$
is the normalized map value, $M_{max}$ and $M_{min}$ denote the
maximum and minimum map intensities respectively. We follow the method
in Kim \textit{et al.} {[}18{]} to measure compactness. Firstly, spatial
clustering is performed on each of the three feature maps F$_{L^{*}}$
, F$_{a^{*}}$ and F$_{b^{*}}$, assuming that a cluster consists
of pixels with similar values and geometrical coordinates. Pixel values
in each feature map F$_{L^{*}}$ , F$_{a^{*}}$ and F$_{b^{*}}$ are
scaled to the range {[}0, 255{]} and quantized to the nearest integers.
Then, for each integer $n$, an observation vector $t_{n}$ is defined
as in Equation \ref{eq:2}.

\begin{equation}
t_{n}=\left[\lambda_{x,n},\lambda_{y,n},\beta n\right]^{T},\;0\;\le\; n\;\text{\ensuremath{\le}}\;255,\label{eq:2}
\end{equation}

\noindent where $\lambda$$_{x,n}$ and $\lambda$$_{y,n}$ denote
the average $x$ and $y$ coordinates of the pixels with value $n$,
and $\beta$ is a constant factor for adjusting the scale of a pixel
value to that of a pixel position. $\beta=\frac{max\{W,H\}}{256}$,
where $W$ and $H$ are the width and height of the input image. These
256 observation vectors are now partitioned into $k$ clusters, \{$R_{1}$,
$R_{2}$, . . . , $R_{k}$\}, using the $k$-means clustering {[}21{]}.
The number $k$ of clusters is eight in this paper. Now, the compactness
$c(R_{k})$ of each cluster $R_{k}$ as defined in Equation \ref{eq:3},
is measured as being inversely proportional to the spatial variance
of pixel positions in $R_{k}$.

\begin{equation}
c(R_{k})=exp(-\alpha.\frac{\sigma_{x,k}+\sigma_{y,k}}{\sqrt{W^{2}+H^{2}}}),\label{eq:3}
\end{equation}
where $\sigma_{x,k}$ and $\sigma_{y,k}$ are the standard deviations
of the x and y coordinates of pixels in $R_{k}$, and $\alpha$ is
empirically set to 10. However, to ignore small outliers, $c(R_{k})$
is set to 0 when the number of pixels in $R_{k}$ is less than 3\%
of the image size. This way we get three compactness maps from feature
maps F$_{L^{*}}$, F$_{a^{*}}$ and F$_{b^{*}}$ respectively. By
taking the square root of the sum of squares of the three compactness
maps we obtain the final compactness map as in Equation \ref{eq:4}
after normalization.

{\scriptsize{}
\begin{equation}
compactMap=\sqrt{compactMap_{L^{*}}^{2}+compactMap_{a^{*}}^{2}+compactMap_{b^{*}}^{2}}\label{eq:4}
\end{equation}
}{\scriptsize \par}

Now a segmented region $r_{i}$ obtained by SLIC superpixel method,
is assigned a compactness value $c_{i}$, which is the average compactness
of the pixels within that region or superpixel.

\subsection{Graph Based Saliency Computation}

This section shows the procedures followed to obtain different graphs
and the associated saliency maps. 

\medskip{}

\subsubsection{Construction of Graph $G_{image}$ from input image}

After obtaining the segmented image regions by SLIC superpixel approach,
we proceed to create a graph $G_{image}$ by considering segmented
image regions as nodes and distance (Euclidean distance and feature
space distance) between the regions as edges of the graph as follows:
The edge weight $w_{combined}^{image}(i,j)$ connecting node $i$
(representing region $r_{i}$) and node $j$ (representing region
$r_{j}$) is taken as the product of combined feature distance of
the considered feature values (intensity or color component values)
represented by weight $w_{feature}^{image}(i,j)$ in Equation \ref{eq:6},
spatial distance (Euclidean distance) between the segmented regions
represented by weight $w_{spatial}^{image}(i,j)$ in Equation \ref{eq:7}
and compactness weight, $w_{compactness}^{image}(i,j)$ in Equation
\ref{eq:8}, which varies according to the compactness of $r_{i}$
and $r_{j}$. We followed our base model GBVS {[}1{]}, to formulate
the combined weight as the product of different weights.

\begin{equation}
I^{image}=[I_{L^{*}},I_{a^{*}},I_{b^{*}}]^{T},\label{eq:5}
\end{equation}

\noindent where I$_{L^{*}}$, I$_{a^{*}}$ and I$_{b^{*}}$are the
normalized feature intensity maps corresponding to L{*}, a{*} and
b{*} components of the image, respectively and $I^{image}$ is a vector
containing these three feature maps. 

\begin{equation}
w_{feature}^{image}(i,j)=\sqrt{\sum_{k=1}^{3}(I_{i,k}^{image}-I_{j,k}^{image})^{2}},\label{eq:6}
\end{equation}

\noindent where $I_{i,k}^{image}$ and $I_{j,k}^{image}$ are the
mean intensity values of the feature channel $k$ ($k$ = 1, 2 and
3 for L{*}, a{*} and b{*} channels respectively) considered for nodes
(superpixels) $i$ and $j$ respectively.

\begin{equation}
w_{spatial}^{image}(i,j)=1-(\frac{\sqrt{(x_{i}-x_{j})^{2}+(y_{i}-y_{j})^{2}}}{D}),\label{eq:7}
\end{equation}

\noindent where $x_{n}$ and $y_{n}$ represent the centroids or the
mean $x$ and $y$ coordinate values of a node $n$ representing a
region $r_{n}$ respectively and $D$ is the diagonal length of the
image.

\begin{equation}
w_{compactness}^{image}(i,j)=(1+\frac{|c_{i}-c_{j}|}{2}),\label{eq:8}
\end{equation}

\noindent where $c_{i}$ and $c_{j}$ represent the compactness of
the regions $r_{i}$ and $r_{j}$ , as explained in the previous section.
The compactness weight factor $w_{compactness}^{image}$ is modeled
as followed in {[}18{]}. This compactness term increases weight $w(i,j)$
, when $r_{i}$ has a low compactness value and $r_{j}$ has a high
compactness value or vice-versa, thus putting more emphasis on the
transition from a less compact object to a more compact object, because
a more compact object is generally regarded as more salient.

{\footnotesize{}
\begin{equation}
w_{combined}^{image}(i,j)=w_{feature}^{image}(i,j).w_{spatial}^{image}(i,j).w_{compactness}^{image}(i,j)\label{eq:9}
\end{equation}
}{\footnotesize \par}

$w_{combined}^{image}$ represents the final edge weight between the
nodes $i$ and $j$.

\medskip{}

\subsubsection{Generation of saliency map \textmd{$M_{GBVS}$ from }$G_{image}$}

We use the graph based visual saliency (GBVS) method in {[}1{]} to
generate a saliency map, $M_{GBVS}$ from the graph $G_{image}$.
Based on the graph structure, we derive an $N\times N$ transition
matrix $TP$, where N is the number of nodes in the graph $G_{image}$.
The element $TP(i,j)$, which is proportional to the graph weight
$w(i,j)$, is the probability with which a random walker at node $i$
transits to node $j$. To obtain $TP$, we first form an $N\times N$
matrix $A$, whose $(i,j)$th element is $A(i,j)=w(i,j)$. The degree
of a node is calculated as the sum of the weights of all outgoing
edges. The degree matrix $W$ of the graph $G_{image}$ is a diagonal
matrix, whose $i$th diagonal element is the degree of node $i$,
as computed in Equation 10.

\begin{equation}
W(i,i)=\sum_{j}w(i,j)\label{eq:10}
\end{equation}

The sum of the elements in each column of $TP$ should be 1, since
the sum of the transition probabilities for a node should be 1. Hence,
we obtain the transition matrix $TP$ as:

\begin{equation}
TP=AW^{-1}.\label{eq:11}
\end{equation}

The movements of the random walker form a Markov chain {[}52{]} with
the transition matrix $TP$. Notice here that, the equilibrium distribution
of Markov chain exists and is unique because the chain is ergodic
(aperiodic, irreducible, and positive recurrent), which can be attributed
to the fact that the underlying graph $G_{image}$ has a finite number
of nodes and is fully connected by construction. The unique equilibrium
(or stationary) distribution $\pi$ of the Markov chain satisfies
Equation 12.

\begin{equation}
\pi=TP\cdot\pi\label{eq:12}
\end{equation}

\begin{figure}
\hspace{1.2cm}\includegraphics[width=10cm]{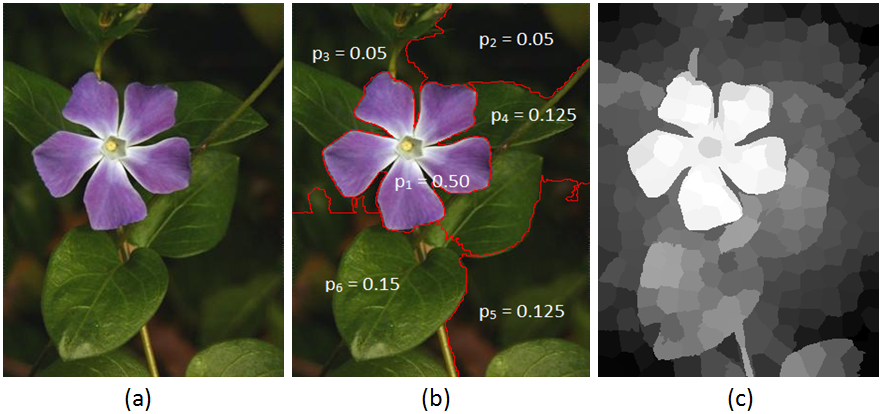}

\caption{\label{fig:2}(a) Original Image (b) Equilibrium distribution probabilities,
$p_{i}$ ($i=1,2,3,4,5,6$) on the segmented image (c) GBVS saliency
map (with 250 segments)}
\end{figure}

The equilibrium distribution of this chain reflects the fraction of
time a random walker would spend at each node/state if he were to
walk forever. In such a distribution, large values are assigned to
nodes that are highly dissimilar to the surrounding nodes. Thus, the
walker at node $i$ moves to node $j$ with a high probability when
the edge weight $w(i,j)$ is large. Transition probabilities (TP)
form an activation measure which is derived from pairwise contrast
in pixel intensities as well as spatial distance between the pixels
{[}1{]}. Here, instead of considering pixels, we group pixels into
superpixels and then consider transition probabilities for the nodes
(each of which represents a superpixel) as being equal to the equilibrium
state probabilities attained on the Markov chain formed on the graph
with edge weight, $w_{combined}^{image}$. Thus, transition probabilities
for all nodes at equilibrium distribution are obtained. A node with
higher equilibrium transition probability represents a more salient
region as compared to another node with lesser probability. Figure
\ref{fig:2} shows how different equilibrium transition probabilities
are assigned to segmented regions (six segments shown for convenience)
and the obtained graph based saliency map. 

Now, let $P(m,n)$ be a pixel ($m$ and $n$ being the pixel coordinates)
which is grouped under a superpixel corresponding to node $i$. Let
$p_{i}=\pi(i)$, where $\pi(i)$ is the $i$th element of the stationary
distribution $\pi$. $\pi(i)$ is the probability that the random
walker stays at node $i$ in the equilibrium condition. Let $p_{max}$
and $p_{min}$ denote the maximum and minimum values of $\pi$ over
all nodes respectively. For each pixel of the image, its saliency
value in the map $M_{GBVS}$ is calculated as in Equation \ref{eq:13}.

\begin{equation}
M_{GBVS}(m,n)=(\frac{p_{i}-p_{min}}{p_{max}-p_{min}})^{2}\label{eq:13}
\end{equation}

In Equation \ref{eq:13}, the map values are obtained by probability
normalization followed by squaring, to highlight conspicuity. This
generates the pixelwise saliency map $M_{GBVS}$ from graph $G_{image}$.
The salient regions are made more salient and the non-salient regions
are adequately suppressed. Thus we get the GBVS saliency map $M_{GBVS}$
by the above method of Markov random walk on the connectivity graph
$G_{image}$. 

\medskip{}

\subsubsection{Construction of Graph $G_{GBVS}$ from \textmd{$M_{GBVS}$}}

\begin{figure}
\hspace{1.2cm}\includegraphics[width=11cm]{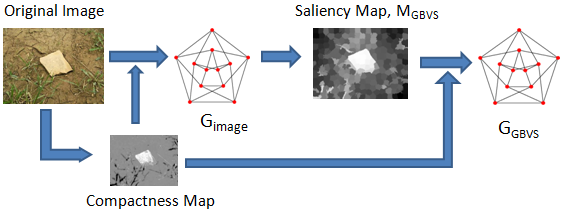}

\caption{\label{fig:3}Construction of graph $G_{image}$(on original image)
and graph $G_{GBVS}$(on saliency map $M_{GBVS}$)}
\end{figure}

Next, the graph $G_{GBVS}$ is constructed based on the saliency map
$M_{GBVS}$. To generate the graph $G_{GBVS}$, we follow a similar
procedure as followed for constructing the graph $G_{image}$. Similar
to the graph $G_{image}$, the graph $G_{GBVS}$ is a fully connected
graph as we consider all possible edges in the graph construction.
The same segmented regions as obtained by SLIC segmentation in case
of $G_{image}$ construction, are considered over the saliency map
$M_{GBVS}$ for creation of graph $G_{GBVS}$. The mean saliency value
of each region in map $M_{GBVS}$ is computed by averaging the saliency
values in the region. $I_{i}{}^{GBVS}$ and $I_{j}{}^{GBVS}$ are
the computed mean saliency values of regions $r_{i}$ and $r_{j}$
respectively, in the map $M_{GBVS}$. The weight $w_{combined}^{GBVS}(i,j)$
of the edge connecting node $i$ and node $j$ (corresponding to regions
$r_{i}$ and $r_{j}$ respectively) is calculated based on spatial
similarity, feature similarity and compactness similarity between
the two segmented regions, as shown in Equation \ref{eq:16}. As the
edge weight $w_{compactness}^{image}$ is calculated based on a separate
clustering procedure as described previously in Equation \ref{eq:8},
the same edge weight which accounts for region compactness is used.
Figure \ref{fig:3} illustrates the followed procedure.

Equations \ref{eq:14} and \ref{eq:15} compute the different edge
weights (feature and spatial respectively) necessary to construct
the graph $G_{GBVS}$.

\begin{equation}
w_{feature}^{GBVS}(i,j)=|I_{i}^{GBVS}-I_{j}^{GBVS}|\label{eq:14}
\end{equation}

\noindent where $I_{i}{}^{GBVS}$ and $I_{j}{}^{GBVS}$ are the mean
saliency values of regions $r_{i}$ and $r_{j}$ respectively, in
the map $M_{GBVS}$.

\begin{equation}
w_{spatial}^{GBVS}(i,j)=1-(\frac{\sqrt{(x_{i}-x_{j})^{2}+(y_{i}-y_{j})^{2}}}{D})\label{eq:15}
\end{equation}

\noindent where $(x_{n},y_{n})$ represents the centroid of a node
$n$ representing a region $r_{n}$ and $D$ is the diagonal length
of the image. $w_{spatial}^{GBVS}(i,j)$ is similar to $w_{spatial}^{image}(i,j)$
in Equation \ref{eq:7}.

\begin{equation}
w_{combined}^{GBVS}(i,j)=w_{feature}^{GBVS}(i,j).w_{spatial}^{GBVS}(i,j).w_{compactness}^{GBVS}(i,j)\label{eq:16}
\end{equation}

\noindent Thus, the graph $G_{GBVS}$ is constructed. The graph $G_{GBVS}$,
based on the saliency map $M_{GBVS}$, is a fully connected graph
or a clique. So, in order to determine the density of this graph to
compute its $k$-dense subgraph, we need to threshold the edges to
form a sparse graph whose weights will be above a certain threshold.
To determine the required threshold for each graph we use the entropy
based thresholding method followed in {[}2{]} .

\subsection{Thresholding the Saliency Graph}

First we select an edge-weight threshold $T$, which is varied between
the minimum and the maximum edge weight in the graph $G_{GBVS}$.
Next taking this threshold $T$, we form two sets of edges, one set
$S_{D}$ representing discarded set of edges and the other set, $S_{S}$
the selected set of edges. Let $w_{i}$ be the weight of an edge $E_{i}$.
For a particular threshold $T$, the ratio of summation of weights
for $S_{D}$ to the weights for the set $S_{D}\cup S_{S}$, $r$ is
calculated as in Equation \ref{eq:17}. 

\begin{equation}
r=\frac{\sum_{w_{i}\leqslant T}w_{i}}{\sum_{i}w_{i}}\label{eq:17}
\end{equation}

Edge-weight entropy $En$ of discarded and selected set of edges is
defined as: 

\begin{equation}
En=-r\log(r)-(1-r)\log(1-r)\label{eq:18}
\end{equation}

\begin{figure}
\includegraphics[scale=0.63]{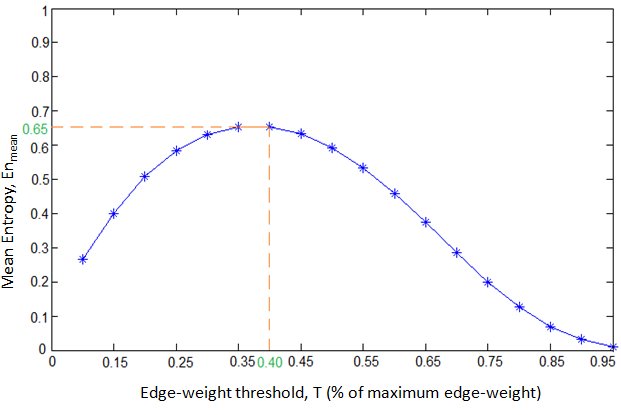}

\caption{\label{fig:4}Variation of mean edge-weight entropy, $En_{mean}$
with edge-weight threshold, \textit{T} (\% of maximum edge-weight)
on the ASD dataset {[}16{]}.}
\end{figure}

Edge-weight entropy $En$ varies with threshold $T$. The threshold
for which edge-weight entropy is maximum is chosen as the edge-weight
threshold $T_{max}$. Note here that, $r$ is a non-decreasing function
of $T$ and \textit{En} attains the maximum value at only one particular
value of \textit{r,} which corresponds to threshold $T_{max}$. Specifically,
$r=0.5+\frac{\sqrt{4+e^{2}}}{2e}$ when $\frac{\partial En}{\partial r}=0$.
Figure \ref{fig:4} shows the variation of the mean entropy, $En$$_{mean}$
of all images in the ASD dataset {[}16{]}. In our experiment, we sampled
the mean entropy value, $En_{mean}$ at a threshold interval of 0.05,
starting from $T_{i}=0.10$ and ending with $T_{f}=0.95$. A threshold\textit{
}value\textit{, }$T_{maxEn}=argmax_{T}(En_{mean})=0.40$ was found
to yield the highest mean entropy, $En_{mean}=0.654$ on the ASD dataset
{[}16{]}. Note here that, $T_{maxEn}=0.40$ shown in Figure \ref{fig:4}
, indicates the threshold value which yields the highest mean entropy
on all images in the ASD dataset {[}16{]}, whereas the threshold $T$
used for an individual image depends on the maximum entropy value
$En$ obtained for that particular image. 

After thresholding the graph $G_{GBVS}$ with threshold $T_{max}$,
we get a modified thresholded sparse graph $G$$_{GBVS}^{thresh}$.
In this paper, we apply dense subgraph computation to a graph to refine
out nodes with high degrees. This ensures that we choose the most
salient regions, as node degree is directly proportional to region
saliency. But a fully connected graph has all nodes with the same
degree. Thus in this case, dense subgraph computation treats all nodes
with equal importance and selects all nodes for inclusion in dense
subgraph set. To circumvent this, we threshold the graph to eliminate
weak edges based on the entropy Equation \ref{eq:18} and allow edges
above a certain threshold value (a threshold that maximizes the entropy
value) to participate in the dense subgraph computation.

\subsection{Dense $k$-Subgraph Computation}

Now we intend to find the $k$-dense subgraph (DkS) from the graph
$G_{GBVS}^{thresh}$ following the procedures used in {[}4{]}. The
density $d_{G}$ of a graph $G(V,E)$ is its average degree. That
is $d_{G}=2|E|/|V|$ . Having discussed about the density of a graph,
we define densest subgraph as a subgraph of maximum density on a given
graph. The objective of the dense $k$-subgraph problem is to find
the maximum density subgraph on exactly $k$ vertices. The problem
is NP-hard, by reduction from Clique. Therefore an approximation algorithm
for the problem is considered. On any input $(G,k)$, the algorithm
returns a subgraph of size $k$ whose average degree is within a factor
of at most $n^{\delta}$ from the optimum solution, where $n$ is
the number of vertices in the input graph $G$, and $\delta<\frac{1}{3}$
is some universal constant. Specifically, for every graph $G$ and
every $1\leq k\leq n$, $A(G,k)\geq\frac{d^{*}(G,k)}{2\cdot n^{1/3}},$
where $A(G,k)$ is the density of the $k$-dense subgraph approximated
with an algorithm $A$ and $d^{*}(G,k)$ is the density of the actual
$k$-dense subgraph in graph $G$. 

\begin{figure}
\hspace{1.2cm}\includegraphics[width=10cm]{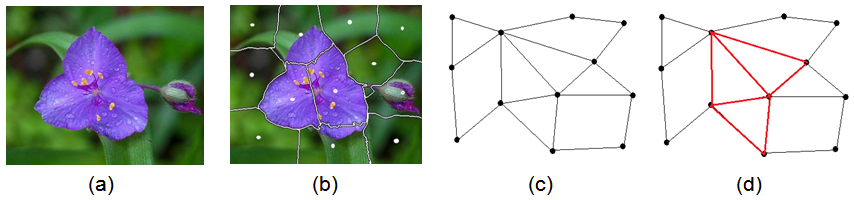}

\caption{\label{fig:5}(a) Original image (b) SLIC segmented image (c) Thresholded
graph (total number of nodes = 12) (d) 6-dense subgraph (red edges)}

\end{figure}

We compute the dense subgraph with $k$ nodes (as defined by the user)
on the thresholded graph we obtained in the previous section, $G_{GBVS}^{thresh}$
following the procedures mentioned in {[}4{]}. The dense $k$-subgraph
problem has an input a graph $G=G(V,E)$ (on $n$ vertices) and a
parameter $k$. The output is $G_{dense}$, a subgraph of $G$ induced
on $k$ vertices, such that $G_{dense}$ is of maximum density and
the density of which is denoted by $d_{G,k}^{*}$. Let us assume that
$G$ has at least $k/2$ edges. Figure \ref{fig:5}(c) shows the sparse
graph formed after thresholding the clique obtained by taking segmented
region centroids (Figure \ref{fig:5}(b)) as nodes, with edge weights
assigned according to methods described in section 3.2. Figure \ref{fig:5}(d)
shows the dense subgraph generated on the segmented regions of the
image. 

We compute the dense subgraph, $G_{dense}$ from the thresholded graph
$G_{GBVS}^{thresh}$ based on the algorithm $A$ (which selects the
best of three different procedures) followed in {[}4{]}. 

The first procedure $A_{1}$ (Procedure 1 in {[}4{]}) selects $k/2$
edges randomly from all edges in the graph $G_{GBVS}^{thresh}$ and
then returns the set of vertices $VS_{1}$ incident with these edges,
adding arbitrary vertices to this set if its size is smaller than
$k$. 

The second procedure $A_{2}$ (Procedure 2 in {[}4{]}) is a greedy
approach which computes a vertex set $VS_{2}$, giving direct preference
to nodes with high degrees. 

The third procedure $A_{3}$ (Procedure 3 in {[}4{]}) first calculates
length-2 walks for all nodes, sorts them and then computes a vertex
set $VS_{3}$ based on the densest subgraph induced on the set $S$
over all nodes. 

Finally, the algorithm outputs the densest of the three subgraphs
(represented by vertex sets $VS_{1}$, $VS_{2}$ and $VS_{3}$) obtained
by the three procedures. Let the densest subgraph obtained be $G_{dense}$
and $S_{dense}$ represent the set of vertices in the densest subgraph,
$G_{dense}$ with $k$ nodes.

\subsection{Final Saliency Map Computation}

Now we compute the map, $M_{dense}$ from the dense subgraph, $G_{dense}$
obtained in the previous step as follows:

\smallskip{}

\textit{Step 1}: Let, set $S_{dense}$ = $\left\{ V_{1},V_{2},...,V_{k}\right\} $
denotes the set of vertices in dense subgraph set, and $M_{dense}(m,n)$
the saliency value at a pixel $P(m,n)$ ($m$ and $n$ being the pixel
coordinates) which is grouped under a superpixel corresponding to
vertex $i$. 

\medskip{}

\textit{Step 2}: For each pixel of the image, the presence of vertex
$i$ corresponding to the pixel, is checked in the set $S_{dense}$.
If found, the degree of vertex $i$, $Deg(Ver_{i})$ is compared with
the mean vertex degree and saliency value assignment is done as showed
in Equation \ref{eq:19}. If vertex $i$ is not found in set $S_{dense}$,
the saliency value at the pixel, $M_{dense}(m,n)$ is assigned a value
zero.

\medskip{}

\textit{Step 3}: The final saliency map, $M_{final}$ is generated
after normalizing the dense subgraph map, $M_{dense}$ according to
Equation \ref{eq:1}, i.e $M_{final}=NormMap(M_{dense}).$

{\scriptsize{}
\begin{equation}
M_{dense}(m,n)=\begin{cases}
(\frac{Deg(Vert_{i})}{max_{\forall i}Deg(Ver_{i})})^{(1/\gamma)} & i\in S_{dense},\\
 & Deg(Ver_{i})>mean_{\forall i}Deg(Ver_{i})\\
\\
(\frac{Deg(Vert_{i})}{max_{\forall i}Deg(Ver_{i})})^{\gamma} & i\in S_{dense},\\
 & Deg(Ver_{i})\leq mean_{\forall i}Deg(Ver_{i})\\
\\
0 & i\notin S_{dense}
\end{cases}\label{eq:19}
\end{equation}
}{\scriptsize \par}

\begin{figure}
\includegraphics[scale=0.62]{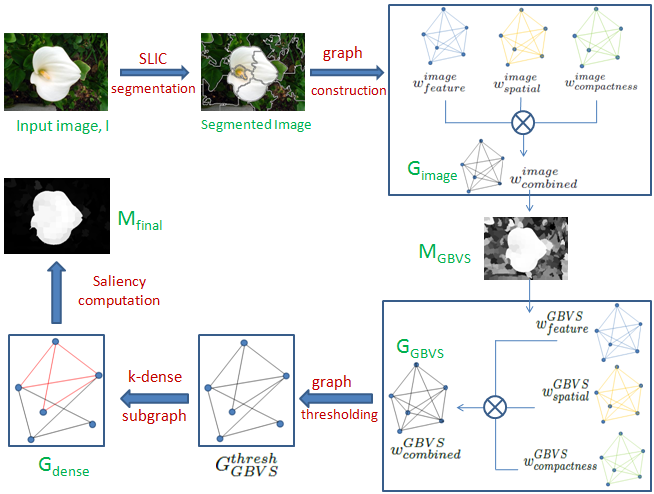}

\caption{\label{fig:6}Detailed steps in the proposed method.}

\end{figure}

\begin{figure}
\hspace{1.6cm}\includegraphics[scale=0.6]{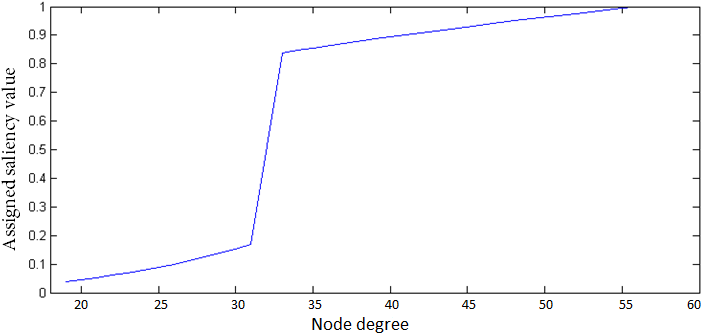}\caption{\label{fig:7}Variation of assigned saliency values of pixels with
varying node degrees. For the generated plot, mean of node degrees
$=30.90$ and $\gamma=3$.}
\end{figure}

From Equation \ref{eq:19}, it may be observed that vertices of the
graph included in the dense subgraph set $S_{dense}$ are given priority
based on their degrees in the subgraph found. The map enhancement
factor, $\gamma$ ($\gamma$ >1) in the final saliency map computation
suppresses the saliency value of pixels in regions that correspond
to nodes with low degrees. On the other hand, pixels corresponding
to nodes with relatively high degrees, closer to the maximum degree
in the dense subgraph set $G_{dense}$ are assigned greater saliency
values. A pixel corresponding to a node (a segmented region) not included
in the dense subgraph set is assigned a value zero. Saliency value
of a segmented region is directly proportional to the corresponding
node degree. Therefore, saliency values of nodes with lower degrees
than the mean degree, which contribute to non-salient regions, are
suppressed and values of nodes with degrees higher than the mean degree,
which contribute to salient regions, are enhanced. This ensures that
sufficient contrast is generated in the saliency map and the salient
regions may be easily distinguished from the non-salient portions.
The variation of saliency values with varying node degrees is shown
in Figure \ref{fig:7}. The final saliency map, $M_{final}$ is obtained
after normalizing the map $M_{dense}$. Figure \ref{fig:6} depicts
the flow of the proposed method in detailed steps. The graphs $G_{image}$
and $G_{GBVS}$ have been shown to be constructed by the multiplication
of the respective feature, spatial and compactness edge-weights.

\subsection{Multiple Salient Region Detection}

\begin{figure}
\hspace{1.3cm}\includegraphics[width=7.9cm,height=6.3cm]{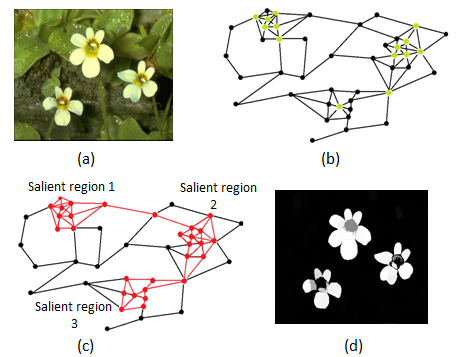}

\caption{\label{fig:8}Multiple salient region extraction using $k$-dense
subgraph algorithm. (a) Original image. (b) Graph, $G_{GBVS}$ (c)
Multiple salient regions extracted after $k$-dense subgraph computation.
(d) Final Saliency map, $M_{final}$.}

\end{figure}

In this section, we demonstrate how multiple salient regions can be
extracted using the proposed method. Figure \ref{fig:8}(b) shows
the graph $G_{GBVS}$ constructed from the input image in Figure \ref{fig:8}(a).
In the example shown, $N=44$ is the total number of graph nodes,
and the value of the parameter $k=28$. The green colored nodes correspond
to nodes with high degrees (degree $\geq$ 5). Procedure 1, which
is a naive method, randomly selects 14 edges ($k/2=14$) and then
includes the set of vertices incident with these edges, adding arbitrary
vertices to this set if its size is smaller than $k$ ($=14$). For
procedure 2, the first fourteen ($k/2=14$) nodes (green colored)
to be included in the dense subgraph set, are selected based on node
degree values. The nodes with the highest number of neighbors in the
already selected set of 14 nodes (marked in green), are the remaining
nodes to be included in the 28-dense subgraph. Procedure 3 followed
in the algorithm may also be analyzed along similar lines. This way,
the vertex sets $VS_{1}$, $VS_{2}$ and $VS_{3}$ are formed from
procedures 1, 2 and 3 respectively. We get the final dense subgraph
(nodes and edges marked in red) in Figure \ref{fig:8}(c) with 28
nodes as the densest among these three sets. This corresponds to three
separate region clusters, which are detected as salient regions by
the algorithm. Figure \ref{fig:8}(d) shows the saliency map obtained
by the dense subgraph computed in Figure \ref{fig:8}(c). It may be
noted here that, the nodes with high degrees are not localized in
image space, as shown in the example. Correspondingly, the dense subgraph
algorithm finds separate dense subgraphs at all image locations, where
the node degrees are high. This property of the algorithm enables
our model to detect multiple salient regions in an image effectively.

\section{Experimental Results and Evaluation}

\subsection{\textit{Datasets}}

We used the following datasets in our experiment.

\begin{itemize}
\item 
Test datasets: 
\begin{itemize} 
\item
\textit{Single salient object dataset}: We used images from the popular MSRA dataset \cite{key-34}, which is the largest object dataset containing 20,000 images in set A and 5,000 images in set B. Achanta \cite{key-20} created a dataset containing 1000 accurate object-contour based human-labeled ground truths corresponding to 1000 images selected from the set B of MSRA salient object dataset. We use the ASD dataset \cite{key-20}, created by Achanta as it enables easy quantitative evaluation.
  To evaluate our method on a slightly more complex dataset, we also use the PASCAL-S dataset for evaluation. The PASCAL-S dataset is derived from the validation set of PASCAL VOC 2010 [54] segmentation challenge and contains 850 natural images surrounded by complex background.

\item
						\vspace{2.5 mm}
\textit{Multiple salient object dataset}:
To demonstrate the efficacy of our model to images with multiple salient objects, we tested the results on the SED2 dataset \cite{key-33}, as it contains 100 images, each with two salient objects. Pixelwise ground truth annotations for salient objects in all 100 images are provided. The CAS model \cite{key-30} was not compared on this dataset, due to lack of author provided results or executable code. The results obtained by running the source codes of the methods (which are made available in their respective websites) or using author provided data, were used for comparison. 
\end{itemize}

\item
Validation dataset: There are two main parameters in the proposed method:
\begin{itemize} 
	\item The $k$ value in $k$-dense subgraph, which selects the participation of nodes in dense subgraph in determining the saliency values of regions, and
	\item The map enhancement factor, $\gamma$, to adjust the quality of saliency maps.
\end{itemize}

To choose these parameters, we used a small validation dataset consisting of 200 images randomly chosen from set A of MSRA dataset \cite{key-34} and pixel accurate salient object labeling obtained from data used in \cite{key-11}. 
		 
\end{itemize}

\subsection{\textit{Experimental Setup}}

\begin{figure}
\hspace{1.3cm}\includegraphics[width=10cm]{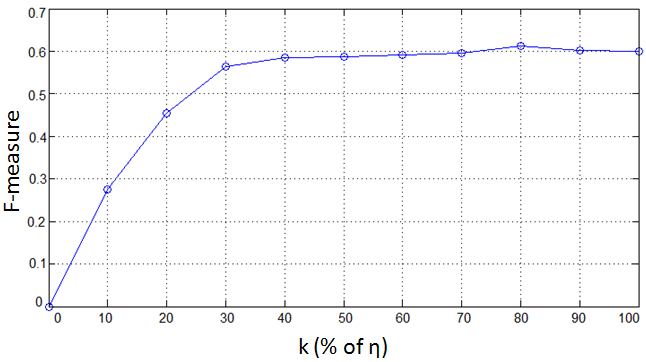}

\caption{\label{fig:9}Plot of F-measure value with $k$-value (used in dense
$k$-subgraph algorithm) on validation dataset. }
\end{figure}

\begin{figure}
\hspace{1.1cm}\includegraphics[scale=0.66]{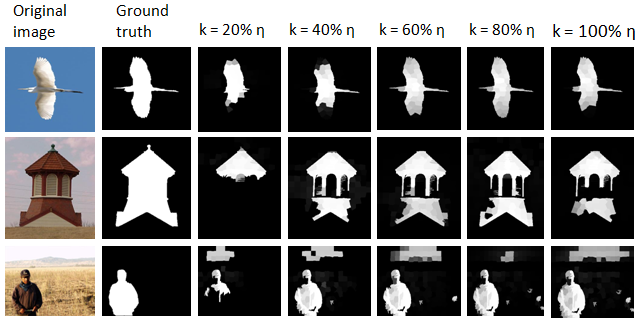}

\caption{\label{fig:11}Saliency maps with varying $k$ ($k$ as in $k$-dense
subgraph) values.}
\end{figure}

We compared our model with eight other well known models, on the test
dataset. These are:

\begin{itemize}
\item Graph based saliency model (GB) \cite{1}(graph based)
\item Frequency tuned saliency model (FT)\cite{key-20}(frequency based)
\item Maximum Symmetric Surround saliency model (MSSS)  \cite{25}(symmetric surround based)
\item Global contrast model (HC, RC)  \cite{key-11}(region based)
\item Over-segmentation model (OS) \cite{key-5}(region segmentation based)
\item Contrast-Aware Saliency model (CAS)  \cite{key-30}(region based)
\item Low rank matrix recovery model (LR)  \cite{key-31}(region based)
\item Simple prior combination model (SDSP)  \cite{key-35}(prior combination based)
\item Principle Component Analysis model (PCA)  \cite{key-53} (region based)
\item Graph based Manifold Ranking model (MR)  \cite{key-36} (graph based)
\end{itemize} We set the number of superpixel nodes, $\eta=250$ for all test images,
as discussed in section 3. As discussed in the previous section, we
used set A of MSRA dataset {[}27{]} as the validation set to choose
the parameters $k$ and the map enhancement factor, $\gamma$. We
calculated the F-measure values (as in Equation \ref{eq:20}) on this
validation set, for $0.1\eta\:\leq\: k\:\leq\:1.0\eta$ ($\eta$ being
the total number of superpixels) and plotted the result as shown in
Figure \ref{fig:9}. $k=0.8\eta$ yielded the highest F-measure value
($F_{\alpha}$ = 0.614). The saliency maps with varying $k$ values
are shown in Figure \ref{fig:11}. It may be observed that, saliency
maps corresponding to $k$ = 80\%$\eta$, resemble the ground truth
data better than other values of $k$. Unwanted image patches appear
for other $k$ values. For lower values of $k$, not all salient regions
get detected and for higher values, unwanted background regions are
labeled as salient regions.

\begin{figure}
\hspace{1.2cm}\includegraphics[scale=0.6]{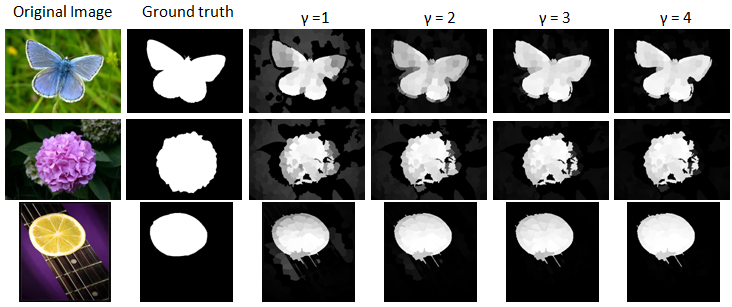}

\caption{\label{fig:10}Effect of varying map enhancement factor, $\gamma$
on final saliency maps, $M_{final}$.}
\end{figure}

Similarly, we calculated the F-measure values for varying map enhancement
factor, $\gamma$ on this dataset, taking $k=0.8\eta$. However, no
significant improvement in F-measure values was observed with increasing
$\gamma$. This is due to the fact that F-measure is based on binarized
maps and non-salient regions with relatively low saliency values are
still assigned zero value in the binarized map, as $\gamma$ value
is decreased. Figure \ref{fig:10} shows the impact of varying the
value of $\gamma$ on the generated saliency maps. We observe that
salient regions become more prominent and stand out from the non-salient
background portions with increasing $\gamma$ value. However saliency
maps cease to improve much with $\gamma>3$. This observation led
us to select $\gamma=3$.

\begin{figure}
\includegraphics[scale=0.64]{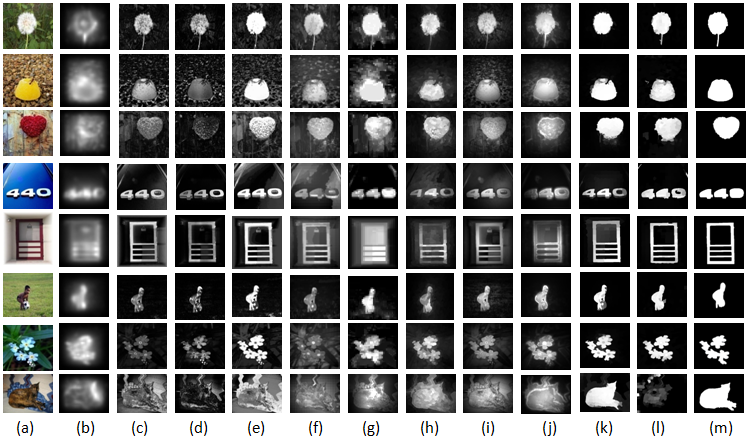}

\caption{\label{fig:12}Comparison of saliency maps obtained by different methods
on the ASD dataset {[}16{]}: (a) Original image, (b) GBVS model {[}1{]},
(c) Frequency tuned model {[}16{]}, (d) Maximum Symmetric Surround
model {[}20{]}, (e) Global contrast model (HC) {[}9{]}, (f) Over-segmentation
model {[}22{]}, (g) Contrast-Aware Saliency model {[}23{]}, (h) Low
rank matrix model {[}24{]}, (i) Simple prior combination model {[}28{]},
(j) Principle component model {[}53{]}, (k) Graph based manifold ranking
model {[}29{]}, (l) Our model and (m) Ground truth.}

\end{figure}

\begin{figure}
\includegraphics[scale=0.59]{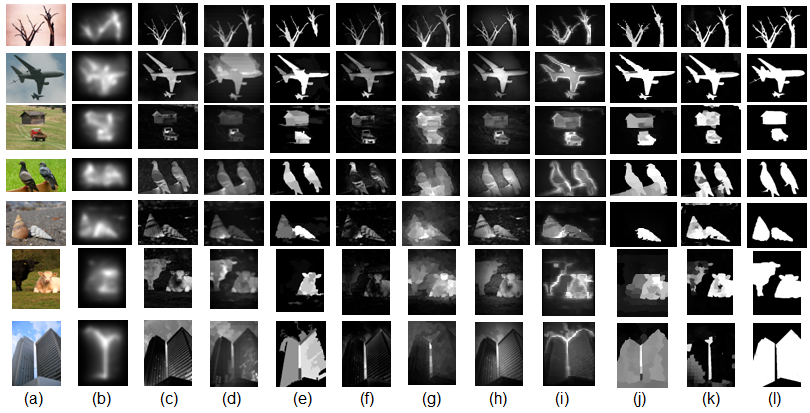}

\caption{\label{fig:13}Comparison of saliency maps obtained by different models
on the SED2 dataset {[}26{]}: (a) Original image, (b) GBVS model {[}1{]},
(c) Frequency tuned model {[}16{]}, (d) Maximum Symmetric Surround
model {[}20{]}, (e) Global contrast model (RC) {[}9{]}, (f) Over-segmentation
model {[}22{]}, (g) Low rank matrix model {[}24{]}, (h) Simple prior
combination model {[}28{]}, (i) Principle component model {[}53{]},
(j) Graph based manifold ranking model {[}29{]}, (k) Our model and
(l) Ground truth.}

\end{figure}

\begin{figure}
\includegraphics[scale=0.58]{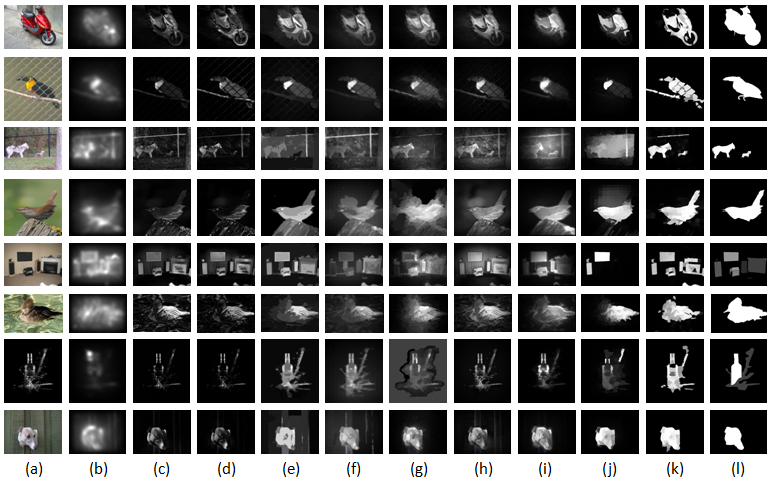}

\caption{\label{fig:14}Comparison of saliency maps obtained by different models
on the PASCAL-S dataset {[}33{]}: (a) Original image, (b) GBVS model
{[}1{]}, (c) Frequency tuned model {[}16{]}, (d) Maximum Symmetric
Surround model {[}20{]}, (e) Global contrast model (RC) {[}9{]}, (f)
Over-segmentation model {[}22{]}, (g) Low rank matrix model {[}24{]},
(h) Simple prior combination model {[}28{]}, (i) Principle component
model {[}53{]}, (j) Graph based manifold ranking model {[}29{]}, (k)
Our model and (l) Ground truth.}

\end{figure}

Figures \ref{fig:12}, \ref{fig:13} and \ref{fig:14} show the qualitative
comparison of results obtained by the proposed method with other well
known models considered in this paper. 

\begin{figure}
\hspace{1.5cm}\includegraphics[width=7.5cm,height=6cm]{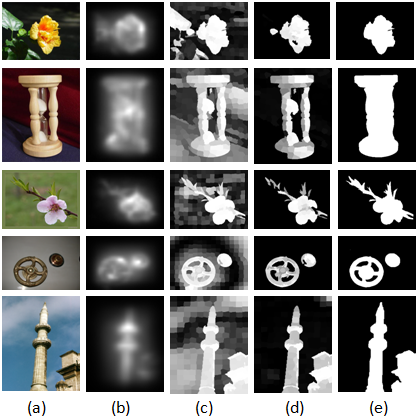}

\caption{\label{fig:15}Comparison of base method (GBVS) and proposed model.
(a) Original image (b) GBVS saliency map (c) Saliency map, $M_{GBVS}$
(d) Final saliency map, $M_{final}$ (e) Ground truth}
\end{figure}

Figure \ref{fig:15} compares the base model, GBVS {[}1{]} with the
two stage saliency maps $M_{GBVS}$ and $M_{final}$ we obtain in
this paper. It may be observed that the blurred saliency maps of the
GBVS algorithm get significant improvement after implementation of
dense subgraph computation. Our model operates on image regions or
superpixels and refines the region based GBVS algorithm as compared
to base model {[}1{]}, which operates at pixel level resulting in
smooth transition from salient to non-salient portions (column (b)).
Region compactness incorporated in the algorithm helps to preserve
object boundaries, overcoming this limitation. The background image
regions which get detected as salient by the region based GBVS algorithm
(maps in column (c)) are eliminated to a great extent in the saliency
maps (column (e)) refined by $k$-dense subgraph algorithm. 

\medskip{}

\subsection{\textit{Evaluation metric} }

The quantitative evaluation of the algorithm is carried out based
on precision, recall, F-measure and Mean Absolute Error. Precision
is a measure of accuracy and is calculated as ratio of number of pixels
jointly predicted salient by binarized saliency map and ground truth
image and the number of pixels predicted salient by the binarized
saliency map. Recall is a measure of completeness and is calculated
as the ratio of number of pixels jointly predicted salient by binarized
saliency map and ground truth and the number of pixels predicted salient
by the ground truth image. F-measure is an overall performance measurement
indicator which is computed as the weighted harmonic mean between
the precision and recall values. It is defined as:

\begin{equation}
F_{\alpha}=\frac{(1+\alpha)\cdot Precision\cdot Recall}{\alpha\cdot Precision+Recall}\label{eq:20}
\end{equation}

\noindent where the coefficient $\alpha$ is set to 1 to indicate
equal importance of precision and recall.

\begin{figure}
\hspace{1.5cm}\includegraphics[width=8.5cm]{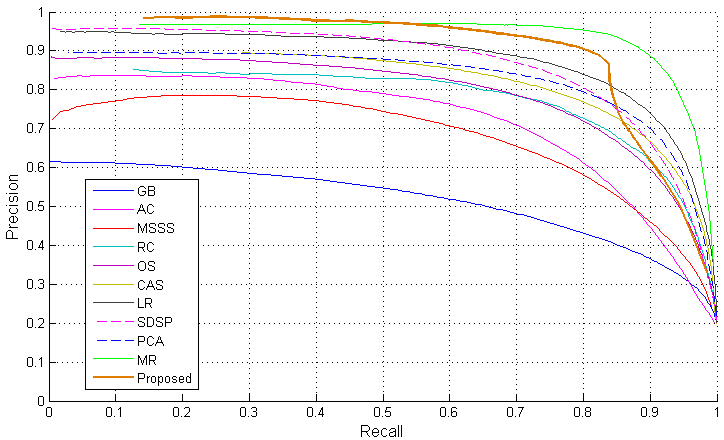}

\hspace{6cm}(a)

\hspace{1.5cm}\includegraphics[width=8.5cm]{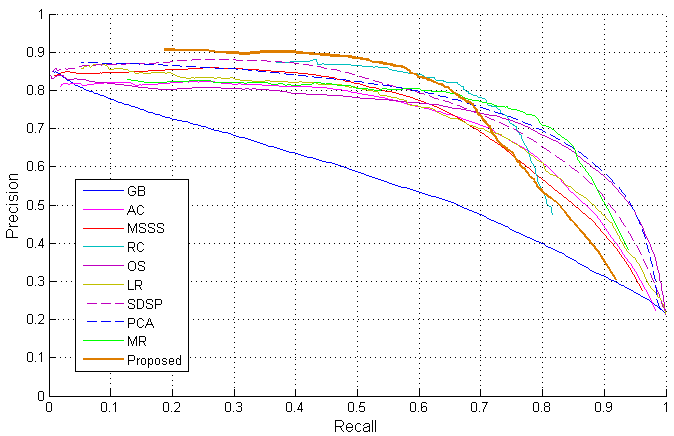}

\hspace{6cm}(b)

\hspace{1.5cm}\includegraphics[width=8.5cm]{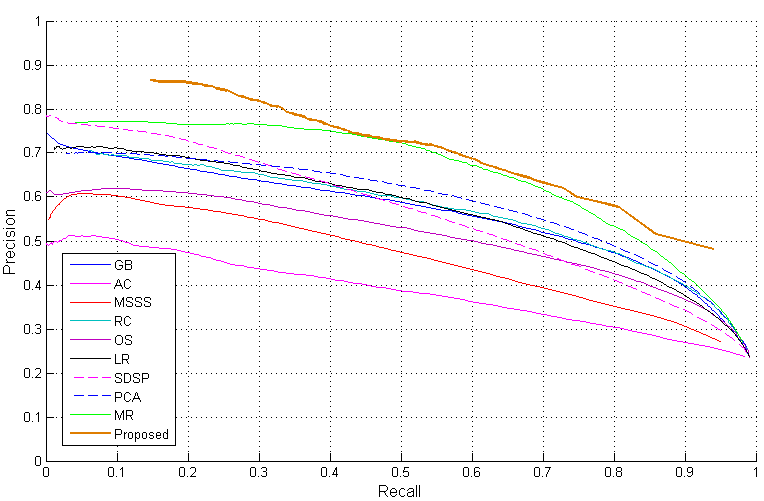}

\hspace{6cm}(c)

\caption{\label{fig:16}Comparison of precision-recall curves of compared models
on (a) ASD dataset {[}16{]} (b) SED2 dataset {[}26{]} (c) PASCAL-S
dataset {[}33{]}. (GB: GBVS method {[}1{]}, FT: Frequency tuned model
{[}16{]}, MSSS: Maximum Symmetric Surround model {[}20{]}, HC, RC:
Global contrast model {[}9{]}, OS: Over-segmentation model {[}22{]},
CAS: Contrast-Aware model {[}23{]}, LR: Low rank matrix model {[}24{]},
SDSP: Simple prior combination model {[}28{]}, PCA: Principle component
model {[}53{]}, MR: Graph based manifold ranking model {[}29{]}, Proposed:
Our method.)}

\end{figure}

\begin{figure}
\hspace{1.5cm}\includegraphics[width=8.4cm]{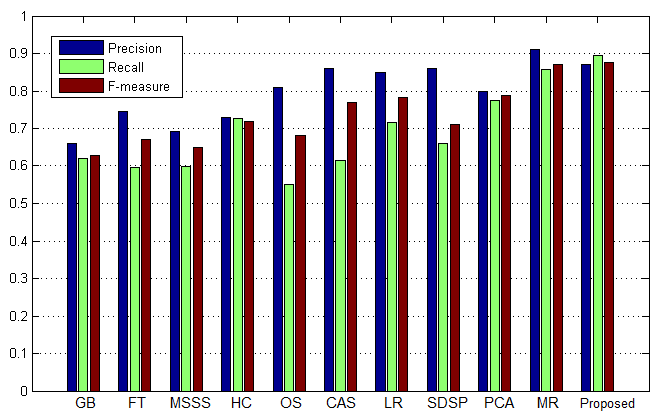}

\hspace{6cm}(a)

\hspace{1.5cm}\includegraphics[width=8.4cm]{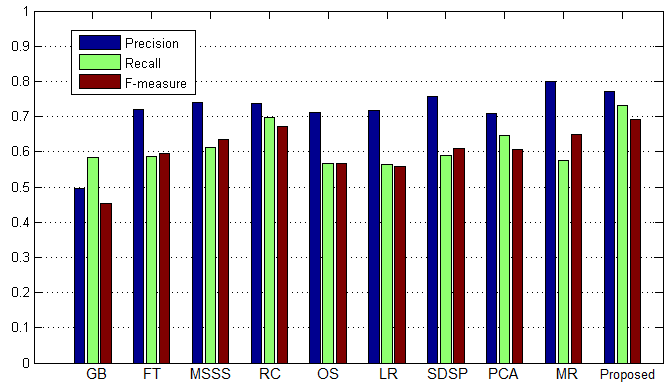}

\hspace{6cm}(b)

\hspace{1.5cm}\includegraphics[width=8.4cm]{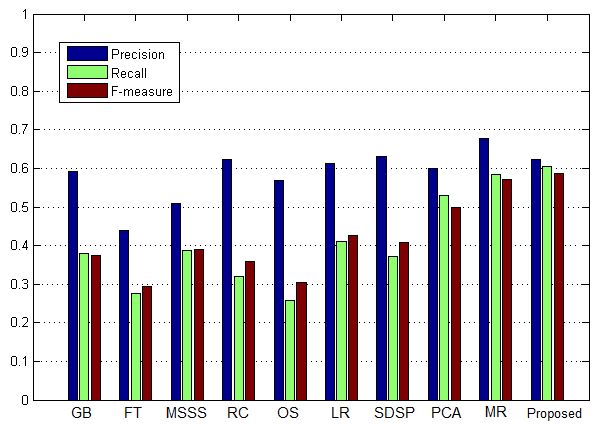}

\hspace{6cm}(c)

\caption{\label{fig:17}Precision, recall, and F-measure value histogram of
the different models on (a) ASD dataset {[}16{]} (b) SED2 dataset
{[}26{]} (c) PASCAL-S dataset. (GB: GBVS method {[}1{]}, FT: Frequency
tuned model {[}16{]}, MSSS: Maximum Symmetric Surround model {[}20{]},
HC, RC: Global contrast model {[}9{]}, OS: Over-segmentation model
{[}22{]}, CAS: Contrast-Aware model {[}23{]}, LR: Low rank matrix
model {[}24{]}, SDSP: Simple prior combination model {[}28{]}, PCA:
Principle component model {[}53{]}, MR: Graph based manifold ranking
model {[}29{]}, Proposed: Our method.)}

\end{figure}

After normalizing the final saliency map, $M_{final}$ to an 8-bit
grayscale image, we threshold the map in the range $T_{f}\in[0,255]$,
to get 256 binarized maps corresponding to each threshold value in
this range. Different precision-recall pairs are obtained for each
of the 256 maps, and a precision-recall curve is drawn. The average
precision-recall curves are generated by averaging the results from
all the 1000 test images from the ASD dataset {[}16{]} (in Figure
\ref{fig:16}(a)), 100 images from the SED2 dataset (in Figure \ref{fig:16}(b))
and 850 images from the PASCAL-S dataset (in Figure \ref{fig:16}(c))
respectively. Furthermore, to evaluate the applicability of saliency
maps for salient object detection more explicitly, we used an image
dependent adaptive threshold ($T_{a}$) to segment objects in the
image, as followed in {[}16{]}. A fixed threshold value in standard
thresholding technique, does not always correctly demarcate the salient
region from the background. Adaptive thresholding overcomes this limitation.
We set the threshold $T_{a}$ as twice the mean saliency value of
the saliency map. Using this adaptive threshold, we obtain the binarized
versions of the saliency maps, for all models. The binarized saliency
maps are then compared to the ground truth images to compute the metrics
of precision, recall, and F-measure for all models compared, as shown
in Figure \ref{fig:17}. These metrics are first computed on all the
test images individually and then averaged over the whole dataset
to obtain the overall performance in terms of average precision-recall
curve and overall F-measure. 

As neither precision nor recall measures consider the number of pixels
correctly marked as non-salient (i.e true negative saliency assignments),
we follow Perazzi \textit{et al.} {[}25{]} to evaluate the Mean Absolute
Error (MAE) for the models compared. MAE between an unbinarized saliency
map S and the binary ground truth G for all image pixels $I_{P}$
is calculated as in Equation \ref{eq:21}. 

\begin{equation}
MAE=\frac{1}{|I|}\sum_{P}|S(I_{P})-G(I_{P})|,\label{eq:21}
\end{equation}

where, $|I|$ is the number of image pixels.

\begin{figure}

\hspace{1.5cm}\includegraphics[width=7.7cm]{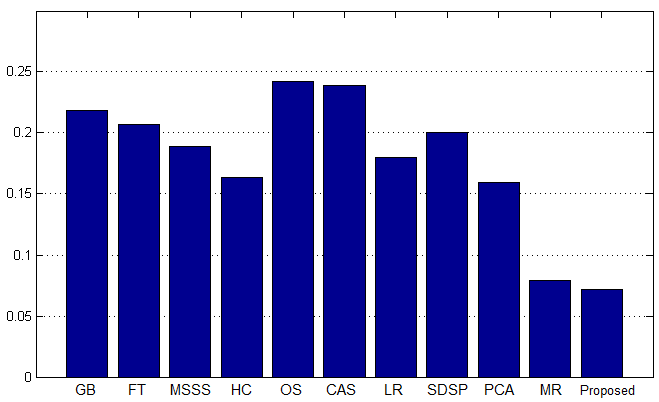}

\hspace{5.5cm}(a)

\hspace{1.5cm}\includegraphics[width=7.7cm]{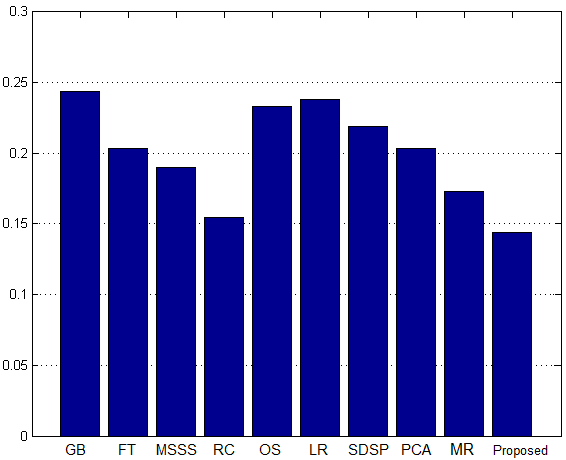}

\hspace{5.5cm}(b)

\hspace{1.5cm}\includegraphics[width=7.7cm]{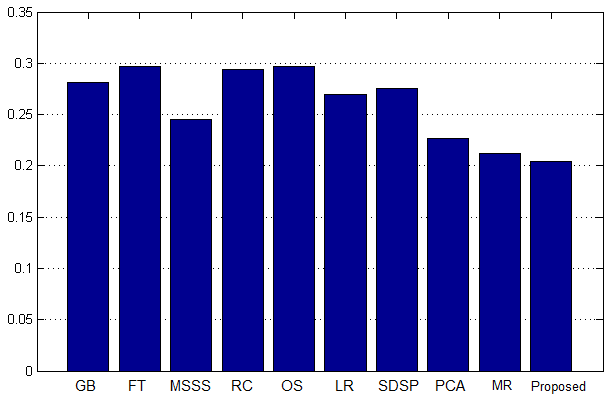}

\hspace{5.5cm}(c)

\caption{\label{fig:18}Average MAE of the different models on (a) ASD dataset
{[}16{]} (b) SED2 dataset {[}26{]} (c) PASCAL-S dataset {[}33{]}.
(GB: GBVS method {[}1{]}, FT: Frequency tuned model {[}16{]}, MSSS:
Maximum Symmetric Surround model {[}20{]}, HC, RC: Global contrast
model {[}9{]}, OS: Over-segmentation model {[}22{]}, CAS: Contrast-Aware
model {[}23{]}, LR: Low rank matrix model {[}24{]}, SDSP: Simple prior
combination model {[}28{]}, PCA: Principle component model {[}53{]},
MR: Graph based manifold ranking model {[}29{]}, Proposed: Our model.)}

\end{figure}

For all methods, we considered the final saliency maps and compared
them to the binary ground truth data to obtain the average MAE on
the used dataset. Results of average MAE evaluation on compared methods
have been shown in Figure \ref{fig:18}.

\subsection{\textit{Evaluation }}

\subsubsection*{Quantitative evaluation}

From Figure \ref{fig:17}, it is clear that the proposed method scores
generally higher precision, recall and F-measure than previously proposed
methods used for comparison. The MR method {[}29{]} scores better
in terms of precision rate as compared to our model on all three datasets,
but our method outperforms it in terms of recall rate and overall
F-measure value. On the PASCAL-S dataset {[}33{]}, the RC {[}20{]}
and the SDSP {[}28{]} methods also have a slight edge over our method
in terms of precision rate, but their recall rates and overall F-measure
values are significantly lower than that of the proposed method. 

It is observed that our algorithm, in general, achieves better recall
values than precision. A high recall value of an algorithm indicates
that most of the relevant results are detected by it. In an experiment
conducted in {[}55{]}, it was found that object region based saliency
models can easily yield high precision values. However, a high recall
value is generally achieved only by conducting the object segmentation
operations either before or after the saliency computation. The high
recall values of the algorithm thus helps our model to yield high
quality object segmentations. Objectness estimation is another major
application of a high recall saliency detection algorithm. Objectness
is generally measured by constructing a small set of bounding boxes
to improve efficiency of the classical sliding window pipeline. High
recall at such a set of bounding box proposals is often a major target.

Average MAE, which provides a better estimate of dissimilarity between
the saliency map and ground truth (as evaluated in Figure \ref{fig:18}),
shows that our method outperforms the existing models by a fair margin,
for all the three datasets. The MR method {[}29{]} performs comparable
to our method on the ASD {[}16{]} and the PASCAL-S {[}33{]} datasets.
On the SED2 dataset {[}26{]} however, average MAE of the RC method
{[}9{]} is most comparable to our method. 

The method proposed is based on the graph based visual saliency model
(GB {[}1{]}) upon which it improvises to extract salient regions using
a graph theoretic model. There is a significant rise in F-measure
value as compared to the GB model (24.5\% i.e from 63.1\% to 87.6\%
on the ASD dataset {[}16{]}, 23.5\% i.e from 45.4\% to 68.9\% on the
SED2 dataset {[}26{]} and 21.4\% i.e from 37.3\% to 58.7\% on the
PASCAL-S dataset {[}33{]}).

\medskip{}

\subsubsection*{Qualitative evaluation}

From the qualitative comparison in Figure \ref{fig:12}, it is observed
that models in columns (e) and (h) yield saliency maps with sufficient
contrast between salient and non-salient regions, however background
regions are still highlighted. On the other hand, the over-segmentation
model {[}22{]} (column (f)) generates low contrast maps and thus not
suitable for object segmentation. The saliency maps generated by the
MR method {[}29{]} generally have nice contrast. However, undesired
background regions are highlighted for some images, such as in rows
5 and 7 or incomplete saliency detection is observed such as in row
4. The proposed model generates saliency maps which are quite similar
to the desired results of salient object segmentation. The salient
regions get more uniformly highlighted with proper suppression of
the background regions, as compared to the other methods. In Figure
\ref{fig:13}, similar observations may be made regarding object shape
retention of multiple objects in our saliency maps, in column (h),
though the method fails for the image in row 7, due to prominence
of background region. The global contrast model {[}9{]} (column (e))
generates comparable maps, however for some images does not highlight
multiple objects as salient, as may be observed in row 6 (only the
right cow is highlighted). Similarly, the MR method {[}29{]} fails
to highlight the left cow (in row 6) and the left shell (in row 5).
For the PASCAL-S dataset {[}33{]} (Figure \ref{fig:14}), our method
clearly highlights the salient regions better than other models. For
instance, in rows 3 and 6, almost all other methods fail to clearly
demarcate the entire salient image region or stress inadequate image
portions as being salient, as in row 2 (only bird neck has more saliency
value). Our saliency model suppresses the non-salient regions effectively
and generates high-resolution saliency maps with well preserved shape
information due to the compactness factor incorporated in the algorithm.
Thus it is inherently advantageous for object segmentation tasks.

\subsection{Computational cost}

In addition to the saliency prediction accuracy, we compare the execution
time of different methods. The computational cost of the compared
methods on a 2.39 GHZ Intel(R) Core i3 CPU with 4GB RAM, are summarized
in Table 1. The software platform was Matlab R2013a. Table 1 shows
the average execution time taken by each saliency detection method
for processing an image on the SED2 dataset {[}26{]}. The computational
costs of different saliency detection methods vary greatly, as seen
from Table 1. The proposed method has lesser execution time than the
LR {[}24{]}, PCA {[}53{]} and CAS {[}23{]} methods. Other methods
run faster than the proposed method, but their saliency detection
accuracies are quite lower than the method proposed as evident from
section 4.4.

\begin{table}
\resizebox{10.2cm}{0.8cm}{%
\begin{tabular}{|c|c|c|c|c|c|c|}
\hline 
Method & GB {[}1{]} & FT {[}16{]} & MSSS {[}20{]} & RC {[}9{]} & OS {[}22{]} & CAS {[}23{]}\tabularnewline
\hline 
\hline 
Time (in sec.) & 1.58 & 0.09 & 0.95 & 0.23 & 3.86 & 6.63\tabularnewline
\hline 
Method & LR {[}24{]} & SDSP {[}28{]} & PCA {[}53{]} & MR {[}29{]} & Proposed & \tabularnewline
\hline 
Time (in sec.) & 25.18 & 0.26 & 11.56 & 0.26 & 6.42 & \tabularnewline
\hline 
\end{tabular}}

\caption{Average execution time of each method on the SED2 dataset {[}26{]}.}
\end{table}

\subsection{Failure Cases and Analysis}

As shown in the previous section, the proposed model outperforms the
compared saliency models on both qualitative and quantitative evaluation.
However, some difficult images are still challenging for the model
proposed as well as other compared models. If an image contains a
part of background regions, which are visually salient against the
major part of background such as row 8 in Figure \ref{fig:12} and
rows 6 and 7 in Figure \ref{fig:13}, the salient object is not properly
highlighted or the nearby background regions are erroneously highlighted
in the generated saliency maps. The proposed model, as well as other
compared saliency models, are yet to be effective to handle such challenging
cases. Also as the proposed algorithm uses intensity, color, and compact
features to determine saliency, it may fail to detect irregular shape
in a visual information scene, since all objects have the same intensity/color
and similar compactness values.

\section{Conclusion and Future Work}

In this article, we have presented a new method for salient region
detection. The proposed method takes the saliency results of the previously
proposed graph based saliency detection method, applies it on segmented
image found by the SLIC superpixel segmentation algorithm and introduces
the $k$-dense subgraph finding problem to that of saliency detection
to improve the extraction of salient parts in a visual information
scene.

Future research scope by this approach may include implementing better
dense subgraph finding algorithms and selection of features used to
construct graph. The method proposed is based on global image features
only. Local image features and contrast information, if considered
in future work, may further enhance the salient region detection ability
of the algorithm proposed. Shape and orientation information can also
be included as a feature to address the issues of irregular shape
detection. We will attempt to incorporate these changes and also generalize
the proposed work in video saliency detection, in our future work.
However, based on the experiments using image data sets labeled with
ground truth salient region, the method followed here has been shown
to provide better region based saliency maps as compared to ten well
known saliency detection methods and is capable of segmenting objects
from an image effectively.

\end{document}